\documentclass[runningheads]{llncs}
\usepackage[utf8]{inputenc}
\usepackage{graphicx}
\usepackage{amsmath,amssymb} 
\usepackage{color}

\usepackage{algorithm}
\usepackage[noend]{algorithmic}
\usepackage{breqn}
\usepackage{subcaption}
\usepackage{gensymb}
\usepackage{siunitx}
\usepackage{xcolor}
\usepackage{cleveref}
\usepackage{float}

\newcommand{\norm}[1]{\left\lVert#1\right\rVert}

\begin{document}
\pagestyle{headings}
\mainmatter
\def\ECCVSubNumber{2724}  

\title{A Closest Point Proposal for MCMC-based Probabilistic Surface Registration} 

\titlerunning{A Closest Point Proposal}
%
\author{Dennis Madsen\thanks{Joint first authors} \and
Andreas Morel-Forster$^\star$ \and Patrick Kahr \and Dana Rahbani \and \\ Thomas Vetter \and Marcel Lüthi}
\authorrunning{D. Madsen et al.}
%
\institute{Department of Mathematics and Computer Science, University of Basel, \\Basel, Switzerland \\
\email{\{dennis.madsen, andreas.forster, patrick.kahr, dana.rahbani, thomas.vetter, marcel.luethi\}@unibas.ch}}
\maketitle

\begin{abstract} 
We propose to view non-rigid surface registration as a probabilistic inference problem. Given a target surface, we estimate the posterior distribution of surface registrations. We demonstrate how the posterior distribution can be used to build shape models that generalize better and show how to visualize the uncertainty in the established correspondence. Furthermore, in a reconstruction task, we show how to estimate the posterior distribution of missing data without assuming a fixed point-to-point correspondence.

We introduce the closest-point proposal for the Metropolis-Hastings algorithm. Our proposal overcomes the limitation of slow convergence compared to a random-walk strategy. As the algorithm decouples inference from modeling the posterior using a propose-and-verify scheme, we show how to choose different distance measures for the likelihood model.

All presented results are fully reproducible using publicly available data and our open-source implementation of the registration framework.
\end{abstract}

\section{Introduction}
The ability to quantify the uncertainty of a surface registration is important in many areas of shape analysis. It is especially useful for the reconstruction of partial data, or for the analysis of data where the exact correspondence is unclear, such as smooth surfaces. Within the medical area, the uncertainty of a partial data reconstruction is needed to make informed surgical decisions \cite{risholm2011probabilistic}. Uncertainty estimates can also be used to build better generalizing Point Distribution Models (PDMs) by assigning an uncertainty measure to each landmark \cite{ma2016weighted,hufnagel2008generation}.{\let\thefootnote\relax\footnote{{Code available at \url{https://github.com/unibas-gravis/icp-proposal}}}}

In this paper, we propose an efficient, fully probabilistic method for surface registration based on Metropolis-Hastings (MH) sampling. Efficiency is gained by introducing a specialized proposal based on finding the closest points between a template and a target mesh.
With this, we benefit from the geometry-aware proposal, while at the same time obtaining an uncertainty estimate for the registration result. 
We formulate the non-rigid surface registration problem as an approximation of the posterior distribution over all possible instances of point-to-point correspondences, given a target surface. With this approach, the registration uncertainty is the remaining variance in the posterior distribution. We use the MH algorithm to sample surface registrations from the posterior distribution. Our method can escape local optima and aims to capture the full posterior distribution of registrations. 

Our method improves on previous works in the literature in different ways. In \cite{risholm2010summarizing,le2016quantifying}, the MH algorithm is used to estimate the uncertainty in non-rigid registration. These papers are working on the image domain, and are not transferable to the surface registration setting. MH has also been used in \cite{morel2018probabilistic} to fit an Active Shape Model to images and in \cite{schonborn2017markov} to fit a Morphable Face Model to an image, both of which only make use of the framework to avoid local optima and to easily integrate different likelihood terms. The main problem with the MH algorithm is the commonly used random-walk approach, which suffers from very long convergence times when working in high-dimensional parameter spaces. To overcome this problem, informed proposal distributions can be designed to perform directed sample updates. In \cite{kortylewski2018informed}, a Bayesian Neural Network is learned to produce informed samples for the MH framework in the case of 3D face reconstruction from a 2D image. This, however, requires training a neural network for each class of shapes to be registered. In \cite{jampani2015informed}, local random-walk is combined with an image-dependent global proposal distribution. This image dependent distribution is, however, not directly transferable to the problem of surface registration. We perform registration by warping a single template mesh to a target surface. No training data is required with our method, whereas most state-of-the-art neural-network-based non-rigid registration methods require thousands of meshes in correspondence for training \cite{deprelle2019learning,groueix20183d}. These methods work well for newer non-rigid registration challenges such as the FAUST dataset \cite{bogo2014faust}, where the focus is to learn shape articulation from a training dataset. This is not something we would advocate using our method for. Our method targets settings with no, or limited, available training data and adds to existing methods by providing an uncertainty measure, which is especially important within the medical domain.

Even though newer registration methods exist, Iterative Closest Point (ICP) and Coherent Point Drift (CPD) are still most commonly used for surface registration scenarios without available training data.\footnote{In this paper, we focus on non-rigid registration. We, therefore, refer to their non-rigid versions whenever ICP or CPD is mentioned.} The ICP algorithm was originally developed for rigid alignment of point sets \cite{chen1992object,besl_method_1992}. ICP iteratively estimates the point-to-point correspondences between two surfaces and then computes a transformation based on the established correspondence. Finally, this transformation is applied to the reference surface. The algorithm has later been modified for non-rigid surface registration \cite{feldmar1996rigid}. In general, standard ICP is very efficient and produces good results. However, a bad initialization may lead the algorithm to local optima from which it is unable to recover. Moreover, it is impossible to estimate the correspondence uncertainty of the registration result. Multiple extensions have been proposed to make the algorithm more robust in scenarios such as missing data or large articulation deformation differences \cite{amberg2007optimal,hufnagel2007point,pan2018iterative}. In \cite{liu2019point}, the ICP method is improved using Simulated Annealing and MCMC. Their method is a robust version of the ICP algorithm, which can find the global optimal rigid registration of point clouds. They do not, however, measure the registration uncertainty, nor are they able to perform non-rigid registrations. An extensive review of different ICP methods can be found in \cite{pomerleau2015review}. The CPD method \cite{myronenko2007non,myronenko2010point} is a probabilistic alternative to ICP. Yet, it does not provide an uncertainty estimate for the registration result.

As it is common in non-rigid registration, we do not infer the rigid alignment. Fortunately, MH allows for easy integration of proposal distributions of parameters other than the shape. The proposal distribution can therefore easily be extended to include translation, rotation, scaling as shown in \cite{morel2018probabilistic}, or texture, illumination, and camera position as in \cite{schonborn2017markov}.

In this paper, we show how closest-point information can be incorporated into the MH algorithm to take geometric information into account. We introduce a novel closest-point-proposal (CP-proposal) to use within the MH algorithm. Our proposal can make informed updates while maintaining the theoretical convergence properties of the MH algorithm. To propose probabilistic surface deformations, we make use of a Gaussian Process Morphable Model (GPMM) as our prior model. In \cite{gerig2018morphable}, GPMMs are applied to face surface registration. However, the authors formulate the registration problem as a parametric registration problem, which does not provide an uncertainty measure for the established correspondence. Several alternatives already exist to the random-walk proposal, such as MALA \cite{grenander1994representations}, Hamiltonian \cite{neal2011mcmc} or NUTS \cite{hoffman2014no}. While the mentioned proposals work well in lower-dimensional spaces and for smoother posteriors, we experienced that they get computationally demanding in high dimensional spaces and have problems when the posterior is far from smooth. 

In our experiments, we register femur bones, where the biggest challenge is establishing correspondence along the long smooth surface of the femur shaft. In another experiment, we use our method to reconstruct missing data. To that end, we compute the posterior distribution of registrations for faces where the nose has been removed. Unlike ICP or CPD, we can give an uncertainty estimate for each point in the surface reconstruction. Furthermore, we show how the standard non-rigid ICP algorithm can end up in local optima, while our method consistently provides good registrations and can also quantify the correspondence uncertainty. The three main contributions of this paper are:
\begin{itemize}
	\item We introduce a theoretically sound, informed closest-point-proposal for the MH framework, with a well-defined transition ratio, \cref{sub:icpProposal}.
	\item We show the usefulness of the posterior distribution of registrations for completing partial shapes and for building better generalizing PDMs, \cref{sec:noseReconstruction}.
	\item We demonstrate that the MH algorithm with our proposal leads to better and more robust registration results than the standard non-rigid ICP and CPD algorithms, \cref{sub:icpVsMCMC}.
\end{itemize}

\section{Background}
In this section, we formally introduce the GPMM as presented in \cite{luthi_gaussian_2018}, and we show how the analytic posterior, which we use in our CP-proposal, is computed.

\subsection{Gaussian Process Morphable Model (GPMM)}\label{sub:gpmmDefinition}
GPMMs are a generalization of the classical point distribution models (PDMs) \cite{luthi_gaussian_2018}. The idea is to model the deformations, which relate a given reference shape to the other shapes within a given shape family, using a Gaussian process. More formally, let $\Gamma_R \subset \mathbb{R}^3$ be a reference surface. We obtain a probabilistic model of possible target surfaces $\Gamma$ by setting
\begin{equation}
\Gamma = \{x + u(x) | x \in \Gamma_R \}
\end{equation}
where the deformation field $u$ is distributed according to a Gaussian process $u \sim GP(\mu, k)$ with mean function $\mu$ and covariance function $k : \Gamma_R \times \Gamma_R \to \mathbb{R}^{3 \times 3}$.
Let $\Gamma_1, \ldots, \Gamma_n$ be a set of surfaces for which correspondence to the reference surface $\Gamma_R$ is known. From this, it follows that we can express any surface $\Gamma_i$ as:
\begin{equation}
\Gamma_i = \{x + u_i(x) | x \in \Gamma_R\}.
\end{equation}
Classical PDMs define the mean and covariance function as:
\begin{align}
\begin{array}{rccl}
\mu_{\text{PDM}}(x) &=& \frac{1}{n} &\sum_{i=1}^n u_i(x)
\\\qquad\\
	k_{\text{PDM}}(x,x') &=& \frac{1}{n-1} &\sum_{i=1}^n (u_i(x)-\mu_{\text{PDM}}(x))  (u_i(x')-\mu_{\text{PDM}}(x'))^T.
\end{array}
\end{align}
However, GPMMs also allow us to define the mean and covariance function analytically. Different choices of covariance functions lead to different well-known deformation models, such as radial basis functions, b-splines, or thin-plate-splines. To model smooth deformations, we choose a zero-mean Gaussian process with the following covariance function: 
\begin{align}
k(x,x') & = g(x,x')*I_{3},
\\
g(x,x') & = s\cdot\exp(\frac{-\norm{x-x'}^2}{\sigma^2}),
\end{align}
where $I_{3}$ is the identity matrix and $g(x,x')$ is a Gaussian kernel.

The model, as stated above, is a possibly infinite-dimensional non-parametric model. In \cite{luthi_gaussian_2018}, they propose to use the truncated Karhunen-Loève expansion to obtain a low-rank approximation of the Gaussian process. In this representation, the Gaussian process $GP(\mu, k)$ is approximated as:
\begin{equation}
u[\vec{\alpha}](x) = \mu(x) + \sum_{i=1}^r \alpha_i \sqrt{\lambda_i} \phi_i(x),\ \alpha_i \sim \mathcal{N}(0, 1)
\end{equation} where $r$ is the number of basis functions used in the approximation and $\lambda_i, \phi_i$ are the i-th eigenvalue and eigenfunction of the covariance operator associated with the covariance function $k$.
Consequently, any deformation $u$ is uniquely determined by a coefficient vector $\vec{\alpha}=(\alpha_1, \ldots, \alpha_r)$. Hence, we can easily express any surface $\Gamma$ as:
\begin{equation}
\Gamma[\vec{\alpha}]= \{x + \mu(x) + \sum_{i=1}^r \alpha_i \sqrt{\lambda_i} \phi_i(x) | x \in \Gamma_R \}
\label{eq:prior_model}
\end{equation}
with associated probability 
\begin{equation}
p(\Gamma[\vec{\alpha}]) = p(\vec{\alpha}) = {(2\pi)}^{-\frac{r}{2}} {\exp(-\left\Vert{\vec{\alpha}}\right\Vert)}^2.
\label{eq:shapeProbability}
\end{equation}

\subsection{Analytical Posterior Model}\label{sec:posteriorModel}
GPMMs make it simple and efficient to constrain a model to match known correspondences, such as user annotations or the estimated correspondence from taking the closest point. Indeed, the corresponding \emph{posterior model} is again a Gaussian process, whose parameters are known in closed form. 

Let $u\sim GP(\mu, k)$ be a GPMM and $\epsilon \sim \mathcal{N}(0,\Sigma)$, $\Sigma=\sigma_{noise} I_{3}$ be the certainty of each known landmark. Every landmark $l_R$ on the reference surface can then be matched with its corresponding landmark $l_T$ on the target. The set $L$ consists of the $n$ reference landmarks and its expected deformation to match the target 
\begin{equation}
L =  \{ (l_{R}^{1}, l_{T}^{1}-l_{R}^{1}), \dots, (l_{R}^{n}, l_{T}^{n}-l_{R}^{n})  \} 	= \{ (l_{R}^{1}, \hat{u}^{1} ), \dots, (l_{R}^{n}, \hat{u}^{n})  \},
\label{eq:posterior-landmarks}
\end{equation}
with $\hat{u}$ being subject to Gaussian noise $\epsilon$.
Using Gaussian process regression, we obtain the posterior model $u_p \sim GP(\mu_{p}, k_{p})$, which models the possible surface deformations that are consistent with the given landmarks. Its mean and covariance function are given by:
\begin{small}
	\begin{align}
		\begin{split}
			\mu_p(x) & = \mu(x) + K_{X}(x)^T(K_{XX}+\epsilon)^{-1}\hat{U} \\
			k_p(x,x') & = k(x,x') + K_{X}(x)^T(K_{XX}+\epsilon)^{-1}K_{X}(x').
			\label{eq:analyticalPosteriorDistribution}
		\end{split}
	\end{align}
\end{small}
Here we defined $K_{X}(x) = \left(k(l_R^i, x)\right)_{i=1,\ldots, n}$, a vector of the target deformation as $\hat{U}=\left(l_T^i-l_R^i\right)_{i=1, \ldots, n}$ and the kernel matrix $K_{XX}=\left(k(l_R^i, l_R^j)\right)_{i,j=1, \ldots, n}$.

\section{Method}
We formulate the registration problem as Bayesian inference, where we obtain the posterior distribution of parameters $\vec{\alpha}$ given the target surface as:
\begin{equation}
P(\vec{\alpha}|\Gamma_{T}) = \frac{P(\Gamma_{T}|\vec{\alpha})P(\vec{\alpha})}{\int P(\Gamma_T | \vec{\alpha})P(\vec{\alpha}) d\vec{\alpha}}.
\label{eq:likelihood-function}
\end{equation}
The prior probability, computed with \cref{eq:shapeProbability}, pushes the solution towards a more likely shape given the GPMM space, by penalizing unlikely shape deformations. The likelihood term can easily be customized with different distance measures and probability functions, depending on the application goal at hand. We are usually interested in modeling the L2 distance between two surfaces ($d_{l^2}$), for which we can use the independent point evaluator likelihood:
\begin{equation}
P(\Gamma_{T}|\vec{\alpha}) = \prod_{i=1}^{n} \mathcal{N}(d_{l^2}(\Gamma_{T}^{i}, \Gamma[\vec{\alpha}]^{i}); 0, \sigma_{l^2}^2),
\label{eq:likelihood}
\end{equation}
as also used in \cite{morel2018probabilistic}. The L2 distance between the $i$-th point $\Gamma[\vec{\alpha}]^{i} \in R^{3}$ and its closest point on the surface $\Gamma_{T}$ is rated using a zero-mean normal distribution with the expected standard deviation for a good registration. The variance $\sigma_{l^2}^{2}$ is the observation noise of the points of our target surface. We can register for a better Hausdorff distance \cite{aspert2002mesh} by changing the likelihood to:
\begin{equation}
P(\Gamma_{T}|\vec{\alpha}) = Exp( d_{H}(\Gamma_{T}, \Gamma[\vec{\alpha}]); \lambda_{H})
\label{eq:likelihood-hausdorff}
\end{equation}
with $d_{H}$ being the Hausdorff distance between the two meshes and $Exp$ being the exponential distribution with pdf ${p(d)=\lambda_{H} e^{-\lambda_{H} d}}$.

\subsection{Approximating the Posterior Distribution}
The posterior distribution defined in \cref{eq:likelihood-function} can unfortunately not be obtained analytically. Yet, we can compute the unnormalized density value for any shape described by $\vec{\alpha}$. This allows us to use the Metropolis-Hastings algorithm \cite{robert2013monte} to generate samples from the posterior distribution in the form of a Markov-chain. The MH algorithm is summarized in \cref{algo:metropolishastings}.
\begin{algorithm}[b]
	\caption{Metropolis-Hastings sampling}
	\label{algo:metropolishastings}
	\begin{algorithmic}[1]
		\STATE $\vec{\alpha}_0 \leftarrow$ arbitrary initialization
		\FOR{$i=0$ to S}
		\STATE $\vec{\alpha}' \leftarrow$ sample from $Q(\vec{\alpha}'|\vec{\alpha}_i)$
		\STATE $t \leftarrow  \dfrac{q(\vec{\alpha}_i|\vec{\alpha}')p(\Gamma_{T}|\vec{\alpha}')p(\vec{\alpha}')}
		{q(\vec{\alpha}'|\vec{\alpha}_i)p(\Gamma_{T}|\vec{\alpha}_i)p(\vec{\alpha}_i)}.$ \COMMENT{acceptance threshold}  \label{algo:sampling:acceptance}
		\STATE $r \leftarrow$ sample from $\mathcal{U}(0,1)$
		\IF{$t > r$}
		\STATE $\vec{\alpha}_{i+1} \leftarrow \vec{\alpha}'$
		\ELSE
		\STATE $\vec{\alpha}_{i+1} \leftarrow \vec{\alpha}_{i}$
		\ENDIF
		\ENDFOR
	\end{algorithmic}
\end{algorithm}
A general way to explore the parameter space is to use a random-walk proposal, i.e. a Gaussian update distribution in the parameter space
\begin{equation}
Q(\vec{\alpha}'|\vec{\alpha}) \sim \mathcal{N}(\vec{\alpha},\sigma_{l}).
\label{eq:proposalDistribution}
\end{equation}
We usually combine differently scaled distributions, each with a specified $\sigma_l$, to allow for both local and global exploration of the posterior distribution. For each proposal, one distribution is chosen at random.

\subsection{CP-proposal}\label{sub:icpProposal}
A random-walk in the parameter space of the prior GPMM model $\mathcal{M}$ is time-consuming as it usually is high-dimensional. Instead, we propose to accelerate convergence by using an informed proposal. For the proposal to reach a unique stationary distribution, we have to be able to compute the transition probability, which requires the proposal to be stochastic. We propose the CP-proposal, which takes the geometry of the model and the target into account to guide the proposed change. Internally, we use a posterior model $\mathcal{M}_{\alpha}$ based on estimated correspondences to propose randomized informed samples. 
From a current state $\vec{\alpha}$, we propose an update $\vec{\alpha}'$ by executing the following steps (visualized in \cref{fig:icp-proposal-visualisation}):
\begin{figure}[t]
	\begin{center}
		\includegraphics[width=0.9\linewidth]{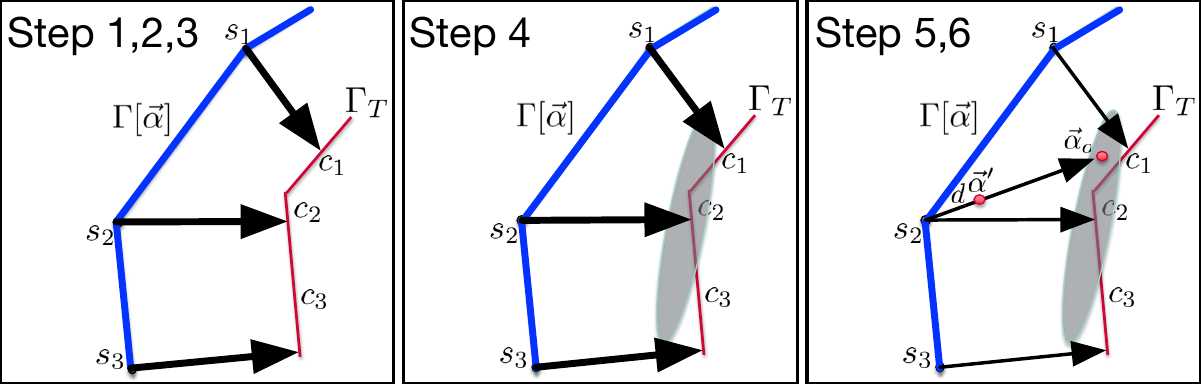}
	\end{center}
	\caption{Visualization of the CP-Proposal with the current instance (\textcolor{blue}{blue}) from the model, and the target surface (\textcolor{red}{red}). The \textcolor{darkgray}{grey} ellipse in the centre window shows the landmark noise for $s_2$. The right window shows how an update for $s_2$ is generated based on the posterior sample $\vec{\alpha}_0$ and the step size $d$.}
	\label{fig:icp-proposal-visualisation}
\end{figure}
\begin{enumerate}
	\item Sample $m$ points $\{s_i\}$ on the current model instance $\Gamma[\vec{\alpha}]$.
	\item For every point $s_i$, $i\in \left[0\ ...\ m\right]$ find the closest point $c_i$ on the target $\Gamma_T$.
	\item Construct the set of observations $L$ based on corresponding landmark pairs ${(s_i,c_i)}$ according to \cref{eq:posterior-landmarks} and define the noise $\epsilon_i\sim \mathcal{N}(0,\Sigma_{s_{i}})$ using \cref{eq:covariance-per-point}.
	\item Compute the analytic posterior $\mathcal{M}_{\alpha}$ (\cref{eq:analyticalPosteriorDistribution}) with $L$ and $\{\Sigma_{s_{i}}\}$.
	\item Get $\vec{\alpha}_{o}$ by first drawing a random shape\footnote{Sampling all ${\alpha}_i$ independently from $\mathcal{N}(0,1)$ and constructing the shape with \cref{eq:prior_model}.} from the posterior model $\mathcal{M}_{ \alpha}$ and then projecting it into the prior model $\mathcal{M}$.
	\item We generate
	\begin{equation}
	\vec{\alpha}' = \vec{\alpha}+d(\vec{\alpha}_{o}-\vec{\alpha})
	\label{eq:linearscaling}
	\end{equation}
	with $d\in[0.0\ ...\ 1.0]$ being a step-length.
\end{enumerate}
The noise $\epsilon$ in step 3 is modeled with low variance along the normal direction and high variance along the surface. The variance at each point $s_i$ in $\Gamma[\vec{\alpha}]$ is computed by:
\begin{equation}
\Sigma_{s_{i}} = [\vec{n},\vec{v}_{1}, \vec{v}_{2}]
\begin{bmatrix}
\sigma_{n}^{2} & 0 & 0 \\ 
0 & \sigma_{v}^{2} & 0 \\
0 & 0 & \sigma_{v}^{2}
\end{bmatrix}
[\vec{n},\vec{v}_{1}, \vec{v}_{2}]^{T},
\label{eq:covariance-per-point}
\end{equation}
where $\vec{n}$ is the surface normal at the position $s_i$ in the mesh and $\vec{v}_{1}$ and $\vec{v}_{2}$ are perpendicular vectors to the normal. The variances along the vectors are set to $\sigma_{n}^{2}=\SI{3.0}{\square\milli\metre}$ and $\sigma_{v}^{2}=\SI{100.0}{\square\milli\metre}$. This noise term ensures that the posterior model from step 4 takes the uncertain correspondence along the surface into account, which is not well defined in flat regions.

If a small step-length is chosen in step 6, the current proposal is only adjusted slightly in the direction of the target surface, resulting in a locally restricted step. With a step size of $1.0$, the proposed sample is an independent sample from the posterior in \cref{eq:analyticalPosteriorDistribution}.
 
In practice, closest-point based updates often find wrong correspondences if the sizes of $\Gamma$ and $\Gamma_T$ greatly differ. This is especially problematic in the case of elongated thin structures. It is, therefore, useful also to establish the correspondence from $\Gamma_T$ to $\Gamma$ from time to time.

\subsubsection{Computing the Transition Probability}
For each new proposal $\vec{\alpha}'$ from the CP-proposal distribution, we need to compute the transition probability as part of the acceptance threshold (see \cref{algo:metropolishastings} step \ref{algo:sampling:acceptance}). The transition probability $q(\vec{\alpha}'|\vec{\alpha})$ is equal to the probability of sampling the shape corresponding to $\vec{\alpha}_{o}$ from the posterior model $\mathcal{M}_{\alpha}$, computed in step 4 of the CP-proposal. For $q(\vec{\alpha}|\vec{\alpha}')$, the transition probability is computed in the same way. We solve \cref{eq:linearscaling} for $\vec{\alpha}'_{o}$ after swapping $\vec\alpha$ and $\vec{\alpha}'$ and evaluate the corresponding shape likelihood under the posterior distribution $\mathcal{M}_{\alpha'}$.

\section{Experiments}
In the following, we perform registration experiments on surfaces of femur bones as well as a reconstruction experiment of missing data on face scans. For the face experiment, the face template and the face GPMM from \cite{gerig2018morphable} are used together with 10 available face scans. For the femur experiments, we use 50 healthy femur meshes extracted from computed tomography (CT) images\footnote{Available via the SICAS Medical Image Repository \cite{kistler2013virtual}.}. Each surface is complete, i.e. no holes or artifacts. This setting is optimal for the standard ICP algorithm and therefore serves as a fair comparison to the CPD algorithm and our probabilistic implementation. The CPD experiments are performed with the MATLAB code from \cite{myronenko2010point} and all other experiments with the Scalismo\footnote{\url{https://scalismo.org}} library.

\subsection{Convergence Comparison}\label{sub:conergenceComparison}
\begin{figure}[t]
	\begin{center}
		\includegraphics[width=\linewidth]{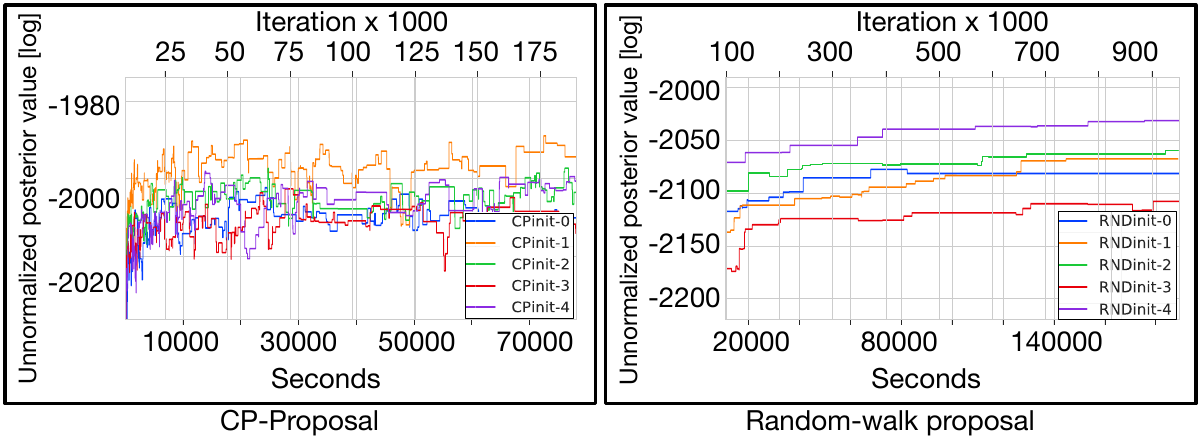}
	\end{center}
	\caption{Posterior plot comparison of the CP-proposal and random-walk. Even with a very small update step for the random-walk, it has difficulties to explore the posterior in the high-dimensional setting. The CP-proposal, on the other hand, can more efficiently explore the high-dimensional space.}
	\label{fig:rndVsIcpPosteriorPlot}
\end{figure}

We compare the convergence properties of our CP-proposal and random-walk. The CP-proposal configuration is mentioned together with its definition in \cref{sub:icpProposal}. For the random-walk, we use a mixture of the proposals defined in \cref{eq:proposalDistribution}, with $\sigma_{l}$ being set to six different levels, from $\SI{1.0}{\milli\metre}$ to $\SI{0.01}{\micro\metre}$, and all six proposal distributions equally likely to be sampled from. 
\begin{figure}[t]
	\begin{center}
		\includegraphics[width=\linewidth]{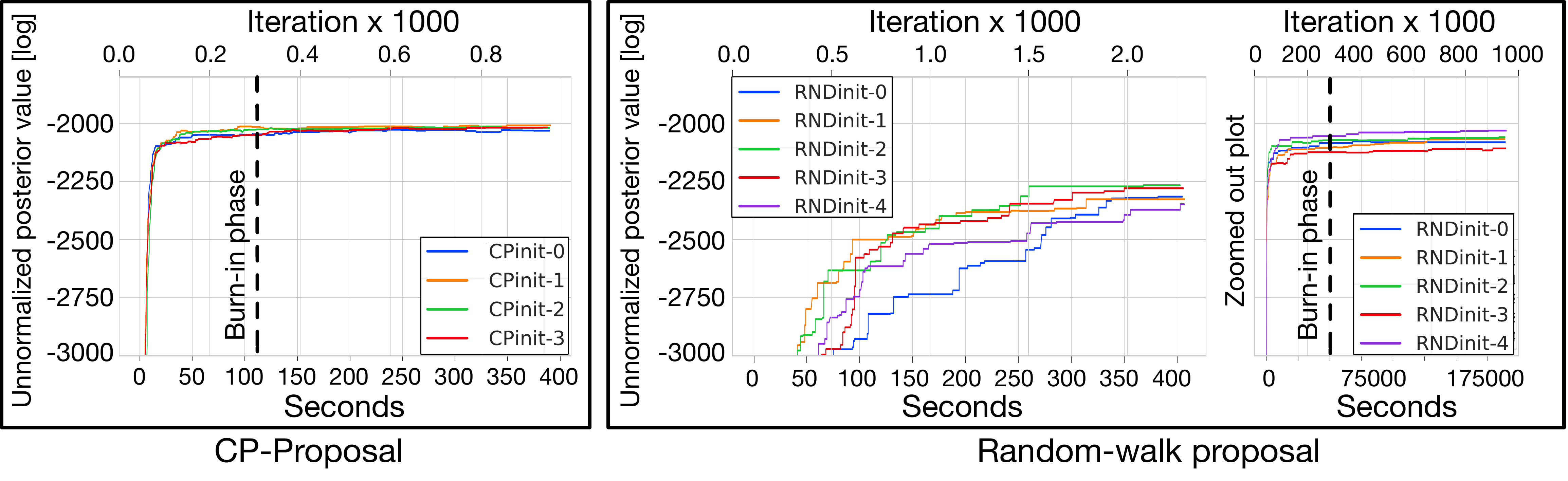}
	\end{center}
	\caption{Convergence plots for the femur GPMM registrations with 50 components. Our CP-proposal is shown to the left and the random-walk (including a zoomed out plot) to the right. The CP-proposal needs 300 iterations, while the random-walk needs more than 200k samples for the burn-in phase without reaching the same registration quality even after 1M iterations. Run-time in seconds is shown on the lower x-axis and number of MH iterations on the upper.}
	\label{fig:rndVsIcpConvergenceTime50components}
\end{figure}
In \cref{fig:rndVsIcpConvergenceTime50components}, the convergence time of the standard random-walk and the CP-proposal is shown. The experiment is performed with a GPMM with a low-rank approximation of 50 components, see \cref{sub:gpmmDefinition}. We randomly initialize the model parameters and start 5 registrations in parallel. As expected, our proposal leads to much faster convergence. In \cref{fig:rndVsIcpPosteriorPlot} we see a posterior plot comparison of the two proposals. Notice how less likely samples are often accepted, which makes it different from deterministic methods such as ICP and CPD. 

\subsection{Posterior Estimation of Missing Data}\label{sec:noseReconstruction}
\begin{figure}[b]
	\begin{center}
		\includegraphics[width=0.95\linewidth]{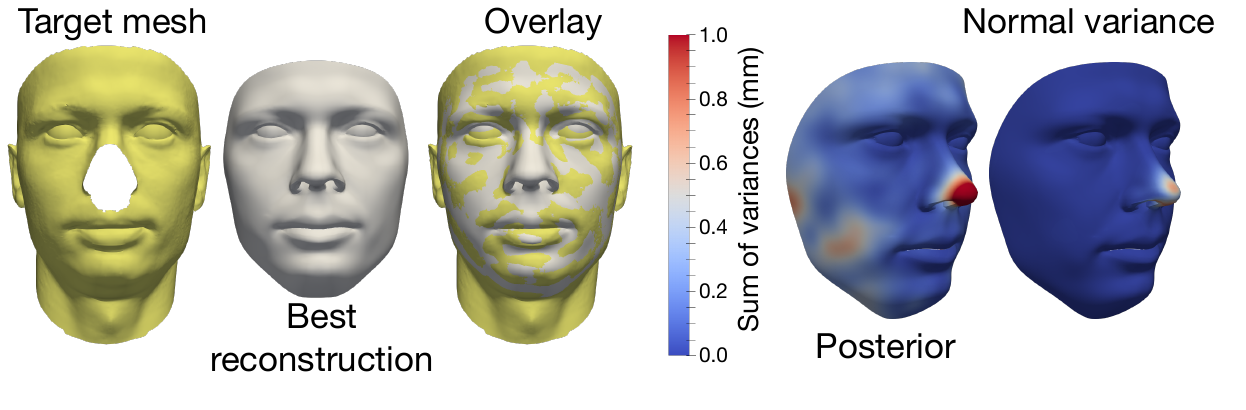}
	\end{center}
	\caption{Nose reconstruction with the face GPMM. No uncertainty in the outer region of the face along the normal indicates that only the correspondence is uncertain compared to the nose.}
	\label{fig:faceFitting00001}
\end{figure}

We use a face GPMM to estimate the posterior distribution of noses from partial face data, where the nose has been removed. In \cref{fig:faceFitting00001}, we see that there is no perfect overlap between the face model and the scan. 
Therefore, we need to change our likelihood function to adjust for a possible changing number of correspondence points during the registration. To obtain a close fit on average, while also penalizing far away points, we use the collective average likelihood introduced in \cite{schonborn2017markov} and extend it with the Hausdorff likelihood,
\begin{dmath}
	P(\Gamma_{T}|\vec{\alpha}) \propto \mathcal{N}(d_{CL}(\Gamma_{T}, \Gamma[\vec{\alpha}]); 0, \sigma_{CL}^2)
	\cdot Exp( d_{H}(\Gamma_{T}, \Gamma[\vec{\alpha}]); \lambda_{H}),
	\label{eq:collectiveAvgLikelihood}
\end{dmath}
where 
\begin{equation}
d_{CL} = \frac{1}{N} \sum_{i=1}^{N}\|\Gamma_{R}^{i}-\Gamma_{T}^{i}\|^{2}.
\end{equation}
Here the closest point of $\Gamma_{R}^{i}$ on the target is $\Gamma_{T}^{i}$ and $N$ is the number of landmarks in the reference mesh $\Gamma_{R}$.

On a technical note, using ICP to predict correspondence in a missing data scenario maps all points from the reference, which are not observed in the target to the closest border. To counter this effect, we filtered away all predicted correspondences, where the predicted target point is part of the target surface's boundary. This is also done in e.g. \cite{amberg2007optimal}.

In \cref{fig:faceFitting00001}, we show the correspondence uncertainty from the posterior registration distribution. Our method infers a larger correspondence uncertainty in the outer region. However, as the surface is observed, the uncertainty is low in the direction of the face surface normals but high within the surface. This is because there is no anatomical distinctive feature to predict the correspondence more precisely. High uncertainty is also inferred on the nose, where data is missing. In contrast to the outer region, the uncertainty in the direction of the normal of the reconstructed nose surface is large. This shows that uncertainty visualization can be used to detect missing areas or to build better PDMs by incorporating the uncertainty of the registration process as demonstrated next. 
\begin{figure}[t]
	\begin{center}
		\begin{subfigure}{0.6\linewidth}
			\includegraphics[width=\linewidth]{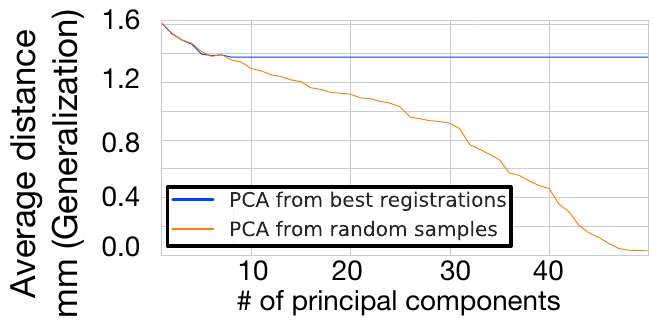}
			\subcaption{}
			\label{fig:generalization}
		\end{subfigure}
		\begin{subfigure}{0.39\linewidth}
		\includegraphics[width=\linewidth]{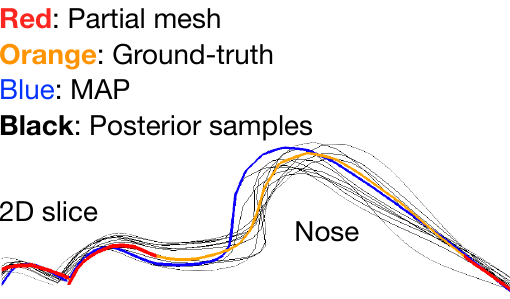}
		\subcaption{}
		\label{fig:nosePosterior-x-slice}
		\end{subfigure}
	\end{center}
	\caption{(a) shows the better generalization of a PDM using 100 random samples from the posterior distributions of each of the 10 targets, compared to only using the most likely samples. The \textcolor{blue}{blue} line flattens out as no more than 9 principal components are computed with 10 meshes. (b) 2D slice view of posterior face samples. The MAP solution does not explain the ground-truth shape, whereas the random samples cover the ground-truth shape.}
	\label{fig:generalization_and_nosePosterior}
\end{figure}
\subsubsection{PDMs from Posterior Distributions}
In this experiment, we show how to benefit from probabilistic registration when building new PDMs from a small dataset with partially missing data, which is common in medical imaging. If much more data is available, the influence of using the posterior is reduced as the variability gained from the probabilistic framework can be gained by including a lot more data. In \cref{sec:noseReconstruction}, we registered the 10 available face scans by computing the posterior distribution of reconstructions for each target. In \cref{fig:nosePosterior-x-slice} we show samples from the posterior distribution of nose reconstructions.

In \cref{fig:generalization}, we compare two different PDMs' generalization abilities, i.e. the capability to represent unseen data \cite{styner2003evaluation}. One PDM is built following the classical approach of only using the most likely registrations. The other PDM is built from 100 random samples from each of the 10 posterior distributions. We compute the average generalization of all 10 PDMs built using a leave-one-out scheme. The plot shows that the PDM built from the posteriors generalizes better.

\subsection{Registration Accuracy - ICP vs CPD vs CP-proposal}\label{sub:icpVsMCMC}
\begin{figure}[t]
	\begin{center}
		\includegraphics[width=0.9\linewidth]{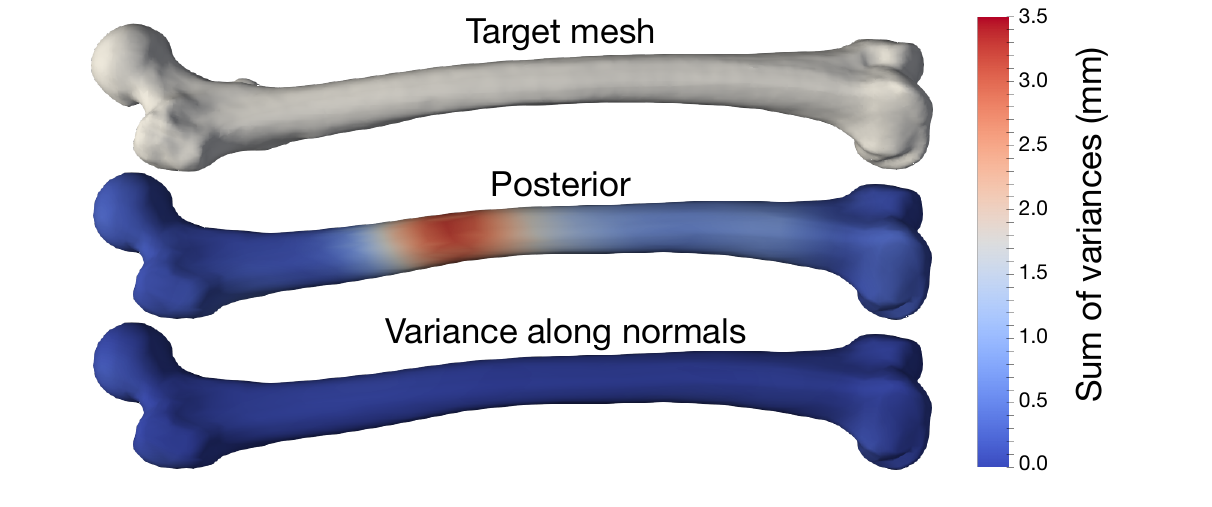}
	\end{center}
	\caption{Femur registration with uncertainty. The registration uncertainty is visualized with the point-wise sum of variances. As expected, the higher uncertainty of the established correspondence coincides with the shaft region with the least characteristic shape. No variance is observed along the normals, so the uncertainty is only in the correspondence along the surface.}
	\label{fig:femurUncertainty}
\end{figure}
\begin{figure}[b]
	\begin{center}
		\includegraphics[width=\linewidth]{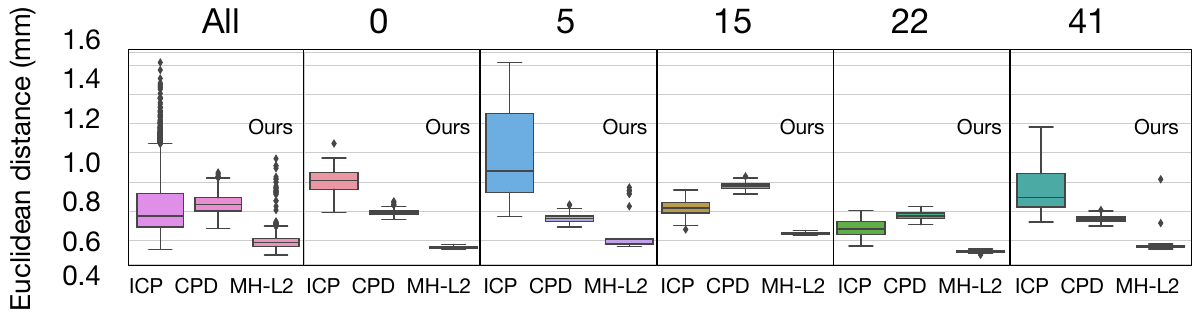}
	\end{center}
	\caption{Distances between the final registrations and their target meshes. For each target mesh, we randomly initialize 100 registrations. The MAP sample from our CP-proposal is superior to ICP and CPD. The femur target id (0 to 49) is shown on top of each plot.}
	\label{fig:icpVsSampling-avgDist}
\end{figure}

In this experiment, we compare the best sample, also known as Maximum a posteriori (MAP) from our probabilistic method, with the ICP and CPD methods. We use a femur GPMM approximated with 200 basis functions. From the model, we sample 100 random meshes as starting points for the registration. We, therefore, end up with 100 registrations for each target. In \cref{fig:icpVsSampling-avgDist} we show the summary of all 5000 registrations (All) and some individual representative target meshes.
The naming scheme combines the registration method (ICP, CPD, or MH with CP) with the likelihood function which was used (L2). The box plot shows the variation of the average L2 surface distances from all 100 registrations of each target. As the ICP and CPD methods converge at maximum 100 iterations, we also restrict our sampling method to 100 samples. The chain usually converges within 100-300 samples as shown in \cref{fig:rndVsIcpConvergenceTime50components}.

The CP-proposal is consistently better than the ICP and CPD registrations, and at the same time provides much less fluctuation in the quality. The few outliers for our method are cases where the chain has not converged. In \cref{fig:femurUncertainty}, we show the uncertainty of the established correspondence of a registration from 1000 samples (300 samples for the burn-in phase). Depicted is the uncertainty of individual points from the posterior distribution from a single registration. Note the high uncertainty values along the shaft, which indicate that the established correspondence is less reliable in that region. No variance along the normals indicates that the uncertainty is purely correspondence shift within the surface.

\subsubsection{Alternative Hausdorff Distance Likelihood}\label{sub:hausdorffLikelihood}
\begin{figure}[t]
	\begin{center}
		\begin{subfigure}{0.49\linewidth}
			\includegraphics[width=2\linewidth]{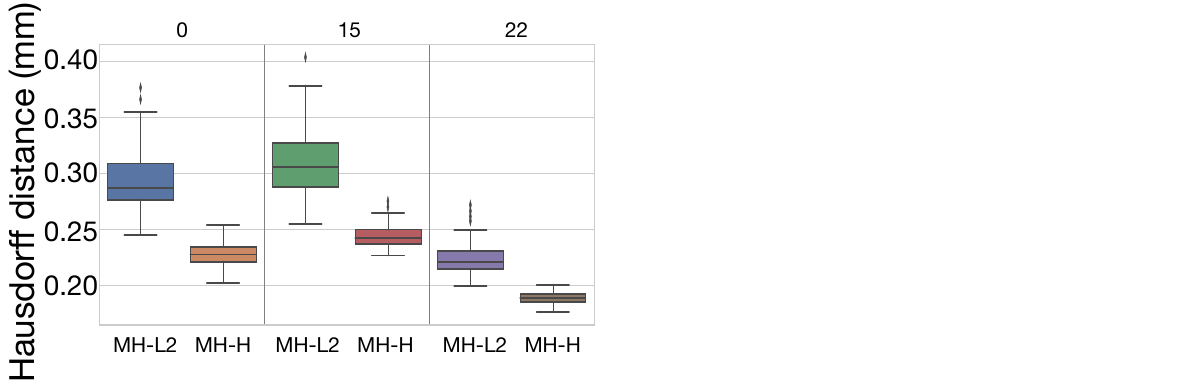}
			\subcaption{}
			\label{fig:icpVsSampling-hausdorffLikelihood}
		\end{subfigure}
		\begin{subfigure}{0.49\linewidth}
			\includegraphics[width=2\linewidth]{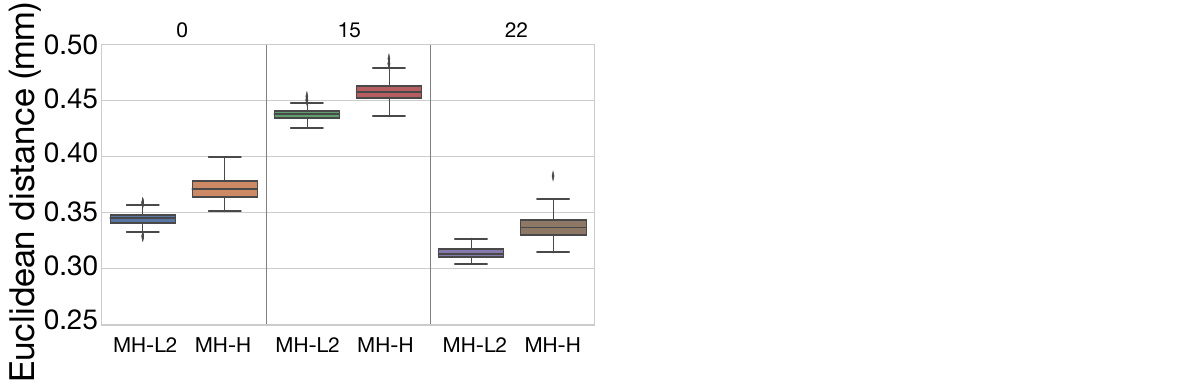}
			\subcaption{}
			\label{fig:icpVsSampling-euclideanLikelihood}
		\end{subfigure}
		\end{center}
	\caption{Registration result comparison using either the Hausdorff or the Euclidean distance likelihood. (a) Euclidean distance and (b) Hausdorff distance between the MAP samples and the target meshes. The plots show that the average L2 surface distance is only slightly worse when the Hausdorff ($H$) likelihood is used.}
\end{figure}
In \cref{fig:icpVsSampling-hausdorffLikelihood}, we compare the registration results based on their Hausdorff distance, and we compare results from sampling using the L2 likelihood, \cref{eq:likelihood}, and the Hausdorff likelihood, \cref{eq:likelihood-hausdorff}. As expected, we can focus the registration to avoid large Hausdorff distances. The equivalent L2 likelihood experiment is shown in \cref{fig:icpVsSampling-euclideanLikelihood} and shows that while optimizing for the Hausdorff distance, the average L2 surface distance is increasing only slightly. This demonstrates the capability to change the likelihood in our framework based on the application's needs.

\subsubsection{Drawbacks of Deterministic Methods}
\begin{figure}[t]
	\begin{center}
		\includegraphics[width=1.0\linewidth]{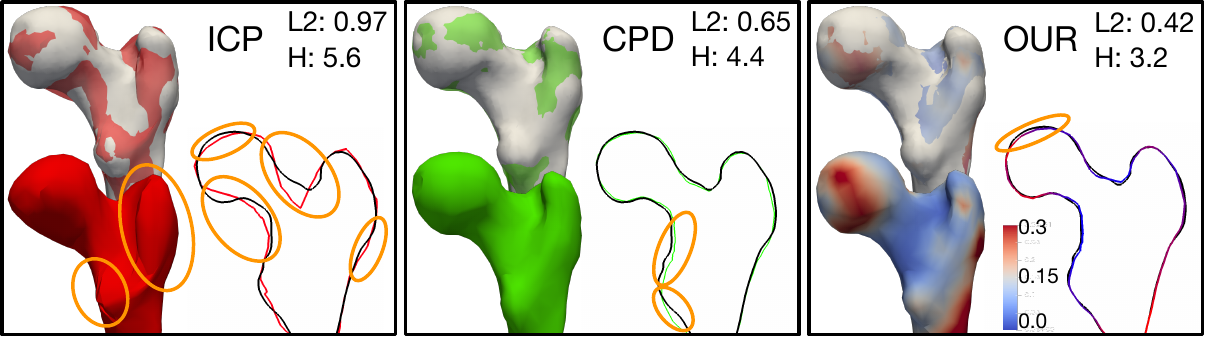}
	\end{center}
	\caption{The 3 registration methods (ICP, CPD, OURS) are shown in separate windows. The registration accuracy for the same target is noted in the form of Euclidean- (L2) and Hausdorff- (H) distances (mm). The \textcolor{orange}{orange} ellipses highlight problematic areas of the registration for each method. For each method, we show the target with the registration overlaid, the 3D-registration, and a 2D slice of the registration (colored) and the target (black). Notice how our method (in comparison to ICP and CPD) shows the correspondence uncertainty (summed point variances (\SI{}{\milli\metre}$^2$) for each landmark).}
	\label{fig:icpFailureCases}
\end{figure}

The main problem with ICP and CPD is that they cannot recover from local optima. If the algorithm finds the closest point on the wrong part of the target, we end up with a bad registration. In \cref{fig:icpFailureCases}, we show a registration result of the 3 registration methods. The ICP method can get the overall length of the bone correct but ends up with a registration, where the structure is folding around itself. The CPD approach is more robust than ICP as it preserves the topological structure of the point sets. Our method additionally provides an uncertainty map.

\subsubsection{Run-time Comparison}
The number of components in the low-rank approximation can be seen as regularization of the deformations. More complex local deformations can be obtained using more components. The algorithm run-time scales linearly in the number of components, with the run-time being 2.5 times slower for each time the model rank doubles. For models with rank 50, 100 and 200, the CP-proposal takes: \SI{46}{\second}, \SI{110}{\second} and \SI{275}{\second}. In comparison, the ICP implementation takes \SI{30}{\second}, \SI{69}{\second} and \SI{155}{\second}. The CPD implementation does not make use of the GPMM model and takes \SI{75}{\second} for 100 samples with our setup. While our method is slower than the two others, we still get reasonable run-times while inferring more accurate results and estimate the full posterior instead of a single estimate.

\section{Conclusion}
In this paper, we presented a probabilistic registration framework. Our main contribution is the informed proposal for the MH algorithm, which makes it possible to work in a high-dimensional model space that would be difficult to explore with pure random-walk. Our informed proposal integrates geometry awareness in the update step, which results in faster convergence. In the case of missing data, our method provides an estimate of the posterior over possible reconstructions. Thus our framework can provide uncertainty measures for critical tasks such as surface reconstruction in the medical domain, as required for surgical decision making. Using our framework, different likelihood terms can be combined and used, while the choice is restricted to the L2 norm in standard ICP and CPD. Finally, we showed how to build PDMs that generalize better using the posterior distribution of registrations.

\subsubsection{Acknowledgements.} This research is sponsored by the Gebert Rüf Foundation
under the project GRS-029/17.
\clearpage

%
%
\bibliographystyle{splncs04}
\bibliography{2724}

\begin{thebibliography}{10}
\providecommand{\url}[1]{\texttt{#1}}
\providecommand{\urlprefix}{URL }
\providecommand{\doi}[1]{https://doi.org/#1}

\bibitem{amberg2007optimal}
Amberg, B., Romdhani, S., Vetter, T.: Optimal step nonrigid icp algorithms for
  surface registration. In: 2007 IEEE Conference on Computer Vision and Pattern
  Recognition. pp.~1--8. IEEE (2007)

\bibitem{aspert2002mesh}
Aspert, N., Santa-Cruz, D., Ebrahimi, T.: Mesh: Measuring errors between
  surfaces using the hausdorff distance. In: Proceedings. IEEE International
  Conference on Multimedia and Expo. vol.~1, pp. 705--708. IEEE (2002)

\bibitem{besl_method_1992}
Besl, P.J., McKay, N.D.: Method for registration of 3-{D} shapes. In: Sensor
  {Fusion} {IV}: {Control} {Paradigms} and {Data} {Structures}. vol.~1611, pp.
  586--607. International Society for Optics and Photonics (1992)

\bibitem{bogo2014faust}
Bogo, F., Romero, J., Loper, M., Black, M.J.: Faust: Dataset and evaluation for
  3d mesh registration. In: Proceedings of the IEEE Conference on Computer
  Vision and Pattern Recognition. pp. 3794--3801 (2014)

\bibitem{chen1992object}
Chen, Y., Medioni, G.: Object modelling by registration of multiple range
  images. Image and vision computing  \textbf{10}(3),  145--155 (1992)

\bibitem{deprelle2019learning}
Deprelle, T., Groueix, T., Fisher, M., Kim, V., Russell, B., Aubry, M.:
  Learning elementary structures for 3d shape generation and matching. In:
  Advances in Neural Information Processing Systems. pp. 7433--7443 (2019)

\bibitem{feldmar1996rigid}
Feldmar, J., Ayache, N.: Rigid, affine and locally affine registration of
  free-form surfaces. International journal of computer vision  \textbf{18}(2),
   99--119 (1996)

\bibitem{gerig2018morphable}
Gerig, T., Morel-Forster, A., Blumer, C., Egger, B., Luthi, M., Sch{\"o}nborn,
  S., Vetter, T.: Morphable face models-an open framework. In: 2018 13th IEEE
  International Conference on Automatic Face \& Gesture Recognition (FG 2018).
  pp. 75--82. IEEE (2018)

\bibitem{grenander1994representations}
Grenander, U., Miller, M.I.: Representations of knowledge in complex systems.
  Journal of the Royal Statistical Society: Series B (Methodological)
  \textbf{56}(4),  549--581 (1994)

\bibitem{groueix20183d}
Groueix, T., Fisher, M., Kim, V.G., Russell, B.C., Aubry, M.: 3d-coded: 3d
  correspondences by deep deformation. In: Proceedings of the European
  Conference on Computer Vision (ECCV). pp. 230--246 (2018)

\bibitem{hoffman2014no}
Hoffman, M.D., Gelman, A.: The no-u-turn sampler: adaptively setting path
  lengths in hamiltonian monte carlo. Journal of Machine Learning Research
  \textbf{15}(1),  1593--1623 (2014)

\bibitem{hufnagel2008generation}
Hufnagel, H., Pennec, X., Ehrhardt, J., Ayache, N., Handels, H.: Generation of
  a statistical shape model with probabilistic point correspondences and the
  expectation maximization-iterative closest point algorithm. International
  journal of computer assisted radiology and surgery  \textbf{2}(5),  265--273
  (2008)

\bibitem{hufnagel2007point}
Hufnagel, H., Pennec, X., Ehrhardt, J., Handels, H., Ayache, N.: Point-based
  statistical shape models with probabilistic correspondences and affine
  em-icp. In: Bildverarbeitung f{\"u}r die Medizin 2007, pp. 434--438. Springer
  (2007)

\bibitem{jampani2015informed}
Jampani, V., Nowozin, S., Loper, M., Gehler, P.V.: The informed sampler: A
  discriminative approach to bayesian inference in generative computer vision
  models. Computer Vision and Image Understanding  \textbf{136},  32--44 (2015)

\bibitem{kistler2013virtual}
Kistler, M., Bonaretti, S., Pfahrer, M., Niklaus, R., B{\"u}chler, P.: The
  virtual skeleton database: an open access repository for biomedical research
  and collaboration. Journal of medical Internet research  \textbf{15}(11),
  ~e245 (2013)

\bibitem{kortylewski2018informed}
Kortylewski, A., Wieser, M., Morel-Forster, A., Wieczorek, A., Parbhoo, S.,
  Roth, V., Vetter, T.: Informed mcmc with bayesian neural networks for facial
  image analysis. arXiv preprint arXiv:1811.07969  (2018)

\bibitem{le2016quantifying}
Le~Folgoc, L., Delingette, H., Criminisi, A., Ayache, N.: Quantifying
  registration uncertainty with sparse bayesian modelling. IEEE transactions on
  medical imaging  \textbf{36}(2),  607--617 (2016)

\bibitem{liu2019point}
Liu, H., Liu, T., Li, Y., Xi, M., Li, T., Wang, Y.: Point cloud registration
  based on mcmc-sa icp algorithm. IEEE Access  (2019)

\bibitem{luthi_gaussian_2018}
L{\"u}thi, M., Gerig, T., Jud, C., Vetter, T.: Gaussian {Process} {Morphable}
  {Models}. IEEE Transactions on Pattern Analysis and Machine Intelligence
  pp.~1--1 (2018)

\bibitem{ma2016weighted}
Ma, J., Lin, F., Honsdorf, J., Lentzen, K., Wesarg, S., Erdt, M.: Weighted
  robust pca for statistical shape modeling. In: International Conference on
  Medical Imaging and Augmented Reality. pp. 343--353. Springer (2016)

\bibitem{morel2018probabilistic}
Morel-Forster, A., Gerig, T., L{\"u}thi, M., Vetter, T.: Probabilistic fitting
  of active shape models. In: International Workshop on Shape in Medical
  Imaging. pp. 137--146. Springer (2018)

\bibitem{myronenko2010point}
Myronenko, A., Song, X.: Point set registration: Coherent point drift. IEEE
  transactions on pattern analysis and machine intelligence  \textbf{32}(12),
  2262--2275 (2010)

\bibitem{myronenko2007non}
Myronenko, A., Song, X., Carreira-Perpin{\'a}n, M.A.: Non-rigid point set
  registration: Coherent point drift. In: Advances in neural information
  processing systems. pp. 1009--1016 (2007)

\bibitem{neal2011mcmc}
Neal, R.M., et~al.: Mcmc using hamiltonian dynamics. Handbook of markov chain
  monte carlo  \textbf{2}(11), ~2 (2011)

\bibitem{pan2018iterative}
Pan, Y., Yang, B., Liang, F., Dong, Z.: Iterative global similarity points: A
  robust coarse-to-fine integration solution for pairwise 3d point cloud
  registration. In: 2018 International Conference on 3D Vision (3DV). pp.
  180--189. IEEE (2018)

\bibitem{pomerleau2015review}
Pomerleau, F., Colas, F., Siegwart, R., et~al.: A review of point cloud
  registration algorithms for mobile robotics. Foundations and
  Trends{\textregistered} in Robotics  \textbf{4}(1),  1--104 (2015)

\bibitem{risholm2011probabilistic}
Risholm, P., Fedorov, A., Pursley, J., Tuncali, K., Cormack, R., Wells, W.M.:
  Probabilistic non-rigid registration of prostate images: modeling and
  quantifying uncertainty. In: 2011 IEEE International Symposium on Biomedical
  Imaging: From Nano to Macro. pp. 553--556. IEEE (2011)

\bibitem{risholm2010summarizing}
Risholm, P., Pieper, S., Samset, E., Wells, W.M.: Summarizing and visualizing
  uncertainty in non-rigid registration. In: International Conference on
  Medical Image Computing and Computer-Assisted Intervention. pp. 554--561.
  Springer (2010)

\bibitem{robert2013monte}
Robert, C., Casella, G.: Monte Carlo statistical methods. Springer Science \&
  Business Media (2013)

\bibitem{schonborn2017markov}
Sch{\"o}nborn, S., Egger, B., Morel-Forster, A., Vetter, T.: Markov chain monte
  carlo for automated face image analysis. International Journal of Computer
  Vision  \textbf{123}(2),  160--183 (2017)

\bibitem{styner2003evaluation}
Styner, M.A., Rajamani, K.T., Nolte, L.P., Zsemlye, G., Sz{\'e}kely, G.,
  Taylor, C.J., Davies, R.H.: Evaluation of 3d correspondence methods for model
  building. In: Biennial International Conference on Information Processing in
  Medical Imaging. pp. 63--75. Springer (2003)

\end{thebibliography}
\end{document}